\documentclass[10pt,twocolumn,letterpaper]{article}

 \usepackage[pagenumbers]{cvpr} 

\usepackage{amsmath,amssymb,mathtools,bm}
\usepackage{booktabs}
\usepackage{graphicx}
\usepackage{xspace}
\usepackage{tikz}
\usepackage{placeins}
\usetikzlibrary{arrows.meta,positioning,calc}

\newcommand{\paper}{L\'evyMatch\xspace}

\newcommand{\N}{\mathcal{N}}
\newcommand{\I}{\mathbf{I}}
\newcommand{\X}{\mathbf{X}}
\newcommand{\Y}{\mathbf{Y}}

\newcommand{\F}{\mathbf{F}}
\newcommand{\U}{\mathcal{U}}
\newcommand{\clip}{\operatorname{clip}}

\definecolor{diffblue}{RGB}{79,129,189}
\definecolor{liftgreen}{RGB}{112,173,71}
\definecolor{sbpurple}{RGB}{112,48,160}
\definecolor{lightgray}{RGB}{245,245,245}

\definecolor{cvprblue}{rgb}{0.21,0.49,0.74}
\usepackage[pagebackref,breaklinks,colorlinks,allcolors=cvprblue]{hyperref}

\def\paperID{*****}
\def\confName{CVPR}
\def\confYear{2027}

\title{Lévy-Driven Correspondence Estimation for Registration}

\author{
	Qianliang Wu$^{1}$ \quad
	Jiaqi Yang$^{2}$ \quad
	Wankou Yang$^{5}$ \quad
	Le Hui$^{2}$ \quad		
	Jin Xie$^{3}$ \quad
	Jian Yang$^{4}$ \quad
	Yaqing Ding$^{5}$ \\
	$^{1}$Nantong University \quad
	$^{2}$Northwestern Polytechnical University \quad
	$^{3}$Nanjing University  \\
	$^{4}$Nanjing University of Science and Technology \quad
	$^{5}$Southeast University
	 \\
}

\begin{document}
\maketitle
\begin{abstract}
Finding reliable point correspondences is difficult when point clouds have low overlap or undergo non-rigid deformation. Iterative refinement can correct uncertain matches, but costly network evaluations limit the number of updates. We present \paper, a L\'evy-driven method that uses random jumps to refine a soft matching matrix. At each step, a network uses the current matching state and geometric information to predict a target matching matrix. A Brownian reference bridge gives an explicit formula for the update toward this target. A Gamma random clock sets the time step for each update. The updated matches provide new geometric feedback for the next target prediction. We further propose a fixed front-loaded Gamma policy that assigns more expected clock time to early updates and less to later ones, without retraining or extra network evaluations. Reordering the same sampled Gamma increments shows that placing larger increments early gives higher accuracy than placing them late. On 4DMatch and 4DLoMatch, our method improves both non-rigid feature matching recall (NFMR) and inlier ratio (IR) over the compared methods. The front-loaded policy achieves 93.09\% NFMR and 92.11\% IR on 4DMatch, and 82.79\% NFMR and 79.07\% IR on 4DLoMatch.
\end{abstract}
\raggedbottom
\clubpenalty=10000
\widowpenalty=10000
\setcounter{dbltopnumber}{5}
\renewcommand{\dbltopfraction}{0.95}
\renewcommand{\textfraction}{0.05}
\begin{figure*}[h]
	\centering
	\includegraphics[width=\textwidth]{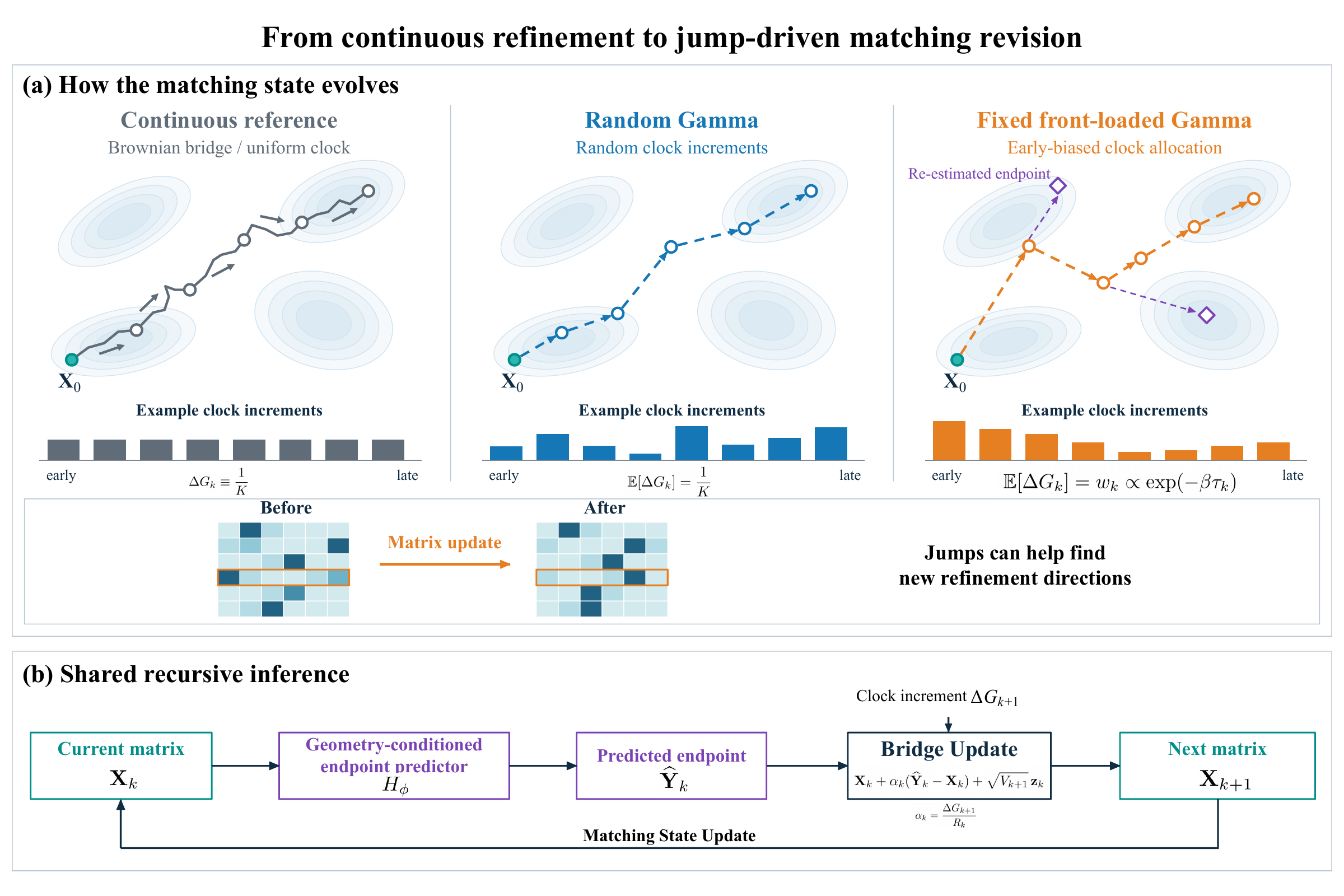}
	\caption{\textbf{Matching-matrix bridge and inference procedure.}
		(a) Example paths for Random Gamma (blue) and Fixed front-loaded Gamma (orange), both starting from the lifted matrix $\X_0$.
		Superscripts $r$ and $f$ identify the two clock policies, respectively, and $h$ denotes the current resolution. Subscript $k$ is the current state index.
		Numbers mark the first three updated states. Diamonds show endpoint predictions $\widehat{\Y}_k$, and squares show final outputs $\widehat{\Y}^{h}$.
		Bars show expected clock increments, not individual samples: $\mathbb{E}[\Delta G_k]=1/K$ for Random Gamma and $\mathbb{E}[\Delta G_k]=w_k$ for Fixed front-loaded Gamma. The normalized weights are defined in Eq.~\eqref{eq:frontloaded_clock}.
		The paths explain the intended effect of early time allocation; they are not measured trajectories.
		(b) The shared endpoint predictor $H_\phi$ estimates $\widehat{\Y}_k$ from the current state and geometric input. A clock-controlled update then gives $\X_{k+1}$ for the next prediction.
		Here $\alpha_k=\Delta G_{k+1}/R_k$, and $R_k=T-U_k$ is the remaining operational time. The variance $V_{k+1}$ is given in Eq.~\eqref{eq:gamma_bridge_transition}, and $\mathbf{z}_k\sim\N(\mathbf{0},\I)$ is newly sampled Gaussian noise.
		Stochastic updates use $k=0,\ldots,K-2$. The last of the $K$ endpoint evaluations returns $\widehat{\Y}^{h}=\widehat{\Y}_{K-1}$ without another stochastic update.}
	\label{fig:framework}
\end{figure*}
\section{Introduction}
\label{sec:intro}
Finding reliable correspondences is a key challenge in registration. Low overlap, repeated structures, sensor noise, and non-rigid deformation can make a match appear locally plausible but globally inconsistent. Selecting such matches too early may discard useful alternatives, while keeping many candidates leaves uncertainty for the alignment stage. Many keypoint-free methods address this problem with coarse-to-fine matching~\cite{yu2021cofinet,qin2022geometric,li2022lepard,yu2023rotation}. Coarse-to-fine matching helps locate candidate regions and distinguish nearby points. However, some correspondences remain uncertain. A soft matching matrix keeps several candidates for each point, allowing their weights to be updated before final matches are selected.

Diffusion-based registration refines matching or alignment estimates through repeated network evaluations~\cite{wu2024diff,chen2023diffusionpcr}. For matching-matrix refinement, a key question is how to reconsider alternatives when the current state favors an incorrect candidate. We investigate whether larger jump-driven changes can help explore these alternatives. By changing the relative weights of candidate matches, a jump may lead the predictor to reassess their geometric consistency and choose a different target matrix. This could provide a more useful direction for later refinement.

To study this idea, we introduce \paper, a Gamma-subordinated bridge in soft matching-matrix space. A Brownian reference bridge defines a continuous stochastic path between an initial matrix and a target, while a Gamma process replaces the uniform clock. Its positive jumps advance the bridge by random amounts of operational time~\cite{hughston2020vginformation,buchmann2016weaksubordination}. Conditioned on the sampled clock and fixed endpoints, the bridge remains Gaussian, giving analytic state distributions and closed-form transitions despite the jump-driven evolution.

The target matrix is unknown at inference, so an endpoint network predicts it from the current state, point features, and geometry. The network predicts where to go, while the bridge determines how to move. As shown in Fig.~\ref{fig:framework}, the predicted endpoint determines the direction of the mean update, while the sampled clock controls the amount of progress and transition variance. After a jump, the updated matching matrix may produce a different alignment and a new endpoint prediction. This feedback allows the network to reassess the update direction from the new matching state.

The timing of these jumps also matters. An early jump leaves more steps to assess alternative matches and correct errors. Later updates can then focus on refining useful correspondences. This motivates allocating more expected operational time to early steps and less to later ones. We consider two clock policies: \emph{Random Gamma} assigns equal expected operational time to each interval, whereas \emph{Fixed front-loaded Gamma} uses decreasing shape parameters to allocate more expected time to early intervals and less to later ones. Both policies preserve the same total-time distribution and use the same trained network. The front-loaded policy changes the distribution of increments at each step while preserving the total-time distribution. It does not force actual matrix changes to decrease over time.

Experiments on 4DMatch and 4DLoMatch study the effects of stochastic timing and early time allocation. Random Gamma and the uniform stochastic clock achieve similar average accuracy. With the same sampled Gamma increments, placing larger increments early improves accuracy over random ordering, while placing them late reduces it. Fixed front-loaded Gamma also improves both recall and inlier ratio over Random Gamma, using the same number of network evaluations. These results support allocating more expected time to early updates for more effective refinement.

Our contributions are summarized as follows:
\begin{itemize}
	\item We introduce a L\'evy-driven bridge for point-cloud correspondence estimation. To the best of our knowledge, this is the first use of a Gamma random clock to advance a Brownian reference bridge in soft matching-matrix space for point-cloud registration. The reference bridge has analytic state distributions and closed-form transitions conditioned on the clock and fixed endpoints.

	\item Within this framework, we combine geometry-conditioned endpoint prediction with Gamma-clock bridge updates. Each updated matching state provides new geometric feedback for the next endpoint prediction, linking the analytic transition to the evolving correspondence estimates.
	
	\item We propose a fixed front-loaded Gamma inference policy that allocates more expected operational time to early updates and less to later ones. Using the same trained checkpoint and network-evaluation budget, we study the effects of clock design, transition noise, initialization, and increment order on 4DMatch and 4DLoMatch.

\end{itemize}

\section{Related Work}
\label{sec:related}

\paragraph{Point cloud registration.}
Learning-based registration estimates correspondences from local features and geometric context. FCGF~\cite{choy2019fully} extracts dense descriptors with sparse 3D convolutions, while D3Feat~\cite{bai2020d3feat} jointly learns keypoint detection and description. For partial overlap, Predator~\cite{huang2021predator} exchanges information between point clouds through overlap attention to identify points likely to yield matches. CoFiNet~\cite{yu2021cofinet} restricts dense matching to patches selected by coarse correspondences. GeoTransformer~\cite{qin2022geometric} encodes pairwise distances and triplet angles to make geometric attention invariant to rigid transformations. REGTR~\cite{yew2022regtr} directly regresses corresponding point positions and overlap probabilities from transformer features. Rotation handling also differs across architectures: RoITr~\cite{yu2023rotation} incorporates point-pair features into rotation-invariant attention, whereas PARE-Net~\cite{pareneteccv24} learns position-aware rotation-equivariant features and uses them to generate a transformation hypothesis from a single correspondence.

Non-rigid matching must accommodate changes in geometry when establishing correspondences. Lepard~\cite{li2022lepard} separates feature and position representations, encodes relative 3D positions, and repositions the point clouds to update cross-cloud positional relations. GraphSCNet~\cite{qin2023deep} exploits local shape preservation through graph-based spatial consistency to reject correspondences that violate local deformation structure. In functional-map-based shape matching, LORD~\cite{xia2026lord} combines local neighborhood constraints with probabilistic deformation refinement to improve pointwise maps. PDCMatch~\cite{xia2026pdcmatch} jointly estimates spectral deformation and correspondence probabilities through expectation maximization, and uses a deformation-based loss for unsupervised learning. SGMatch~\cite{ye2026sgmatch} combines semantic and geometric features through local cross-attention and introduces conditional flow matching to regularize feature transport along interpolated paths. Its flow acts on feature representations rather than on the matching-matrix state.

\paragraph{Correspondence selection and geometric consistency.}
Correspondence selection uses geometric relations to distinguish inliers from false matches. PointDSC~\cite{bai2021pointdsc} combines spatial-consistency-guided non-local feature aggregation with differentiable spectral matching to estimate inlier confidence. SC$^2$-PCR~\cite{chen2022sc2} introduces second-order spatial compatibility, using shared compatible correspondences to reduce the ambiguity of pairwise distance agreement. VBReg~\cite{Jiang_2023_CVPR} models non-local correspondence features through variational Bayesian inference and uses voting to identify inliers. Mutual Voting~\cite{yang2024mutual} jointly updates voter reliability and candidate confidence in a compatibility graph. MAC~\cite{zhang20233dmac} forms consensus sets from maximal cliques to generate multiple transformation hypotheses; MAC++~\cite{zhang2025macplus} adds voting-guided clique pools and progressive hypothesis evaluation. HyperGCT~\cite{zhang2025hypergct} learns higher-order geometric relations with dynamic hypergraphs for hypothesis generation. These approaches operate on candidate correspondences, whereas our refinement updates soft assignment scores before final correspondence extraction.

\paragraph{Diffusion and transport for correspondence refinement.}
Iterative registration models differ in the state they refine. The original DiffusionPCR formulation~\cite{chen2023diffusionpcr} treats the rigid transformation predicted by an existing registration model as a noisy estimate and learns its refinement toward the ground truth. It uses spherical interpolation for rotations and encodes the preceding transformation as conditioning information. FUSER~\cite{jiang2026fuser} jointly estimates global poses from multiple scans; its FUSER-DF variant performs diffusion refinement in the joint $\mathrm{SE}(3)^N$ pose space. Diff-Reg~\cite{wu2024diff} instead diffuses correspondence assignments, learning to denoise Gaussian-perturbed matching matrices in doubly stochastic matrix space. Our method also refines matrix-valued states, but derives updates from conditional reference-bridge transitions under a Gamma random clock and studies the allocation of operational time across a fixed number of evaluations.

Transport formulations provide the path constructions underlying such iterative models. Flow Matching~\cite{lipman2023flow} regresses vector fields along prescribed conditional probability paths without simulating the learned dynamics during training. Stochastic interpolants~\cite{albergo2023building} construct paths between endpoint samples and learn the associated transport velocity. Diffusion Schr\"odinger Bridge~\cite{debortoli2021diffusion} uses iterative proportional fitting to connect endpoint distributions, while DSBM~\cite{shi2023dsbm} introduces iterative Markovian fitting and regression-based bridge learning. For point cloud denoising, P2P-Bridge~\cite{vogel2024p2pbridge} learns a diffusion bridge directly between noisy and clean point sets rather than using a pure-noise source distribution.

\paragraph{L\'evy and jump-driven generative models.}
L\'evy flights have also been used in registration: Zhang et al.~\cite{zhang2024lowoverlap} use them in a cuckoo-search algorithm for rotation-parameter optimization. Our method uses a Gamma random clock to evolve soft correspondence matrices. Jump-driven models also extend generative learning beyond Gaussian diffusion. The L\'evy--It\^o Model (LIM)~\cite{yoon2023levy} uses isotropic $\alpha$-stable L\'evy noise and fractional denoising score matching for image generation. Generator Matching~\cite{holderrieth2025generator} generalizes vector-field and score learning to Markov generators, accommodating continuous dynamics, jumps, and their combinations. Zlotchevski and Chen~\cite{zlotchevski2024jumpdiffusion} characterize Schr\"odinger bridges with jump-diffusion reference measures through $h$-transforms. Gamma subordination and variance-Gamma bridges have established theoretical foundations~\cite{hughston2020vginformation,buchmann2016weaksubordination}. We use a Gamma clock to change the operational time of a Brownian reference bridge in correspondence-matrix space, with endpoints predicted from the evolving state. Our focus is the resulting finite-step refinement and its time allocation.

\section{Background: From Brownian to Random Gamma Bridges}
\label{sec:background}

\subsection{Brownian Matrix Bridge}
Gaussian distributions below act on the vectorized valid matrix entries; padded entries are excluded. We use $t$ for continuous time and $k$ for discrete inference steps.
A Brownian matrix bridge defines a stochastic path from an initial matrix $\X_0$ to a target matching matrix $\Y$. The construction is conditional on the chosen $\X_0$ and permits either a structured or a randomly sampled source. A \emph{clock} controls the amount of internal, or operational, time used by each update. Operational time measures progress along the bridge, not computation time. At a fixed time $t\in[0,1]$, we can sample the Brownian bridge state as
\begin{equation}
    \X_t=(1-t)\X_0+t\Y
    +\sigma_B\sqrt{t(1-t)}\bm{\epsilon},
    \qquad \bm{\epsilon}\sim\N(\mathbf{0},\I).
    \label{eq:brownian_path_background}
\end{equation}
The mean moves linearly from $\X_0$ to $\Y$, and $\sigma_B$ sets the Gaussian noise scale. The noise is zero at both endpoints, and the reference bridge has continuous sample paths. A uniform clock divides operational time equally across updates. Equal time increments, however, do not always produce equal matrix changes. We use this bridge as the reference for our random-clock construction.

\subsection{From L\'evy Processes to a Gamma Clock}

A L\'evy process starts at zero, has independent and stationary increments, and is continuous in probability~\cite{applebaum2009levy}. Independent increments mean that changes over non-overlapping intervals are independent. Stationary increments mean that their distributions depend only on the interval length. Brownian motion is the continuous Gaussian case. A Gamma process moves forward through positive jumps. Its accumulated value never decreases, so it can act as a random clock.

An independent Gamma clock $G_t$ turns Brownian motion $B_t$ into the variance-Gamma process $B_{G_t}$~\cite{buchmann2016weaksubordination}. In other words, we observe the Brownian process at times given by the Gamma clock. A clock jump advances the process by a finite amount of operational time. We apply this time change to a Brownian matrix bridge over the sampled total time~\cite{hughston2020vginformation}. Given the clock and fixed endpoints, the reference bridge still has a Gaussian form.

\subsection{Random Gamma Matrix Bridge}
\label{sec:bg_unified}

Let $G_t$ be a Gamma subordinator starting at $G_0=0$, with independent increments
\begin{equation}
    G_t-G_r\sim\operatorname{Gamma}\!\left(\kappa(t-r),
    \operatorname{rate}=\kappa\right),\qquad r<t,
    \label{eq:gamma_subordinator}
\end{equation}
Here $\operatorname{Gamma}(a,\operatorname{rate}=b)$ has mean $a/b$ and variance $a/b^2$. The concentration parameter $\kappa>0$ controls variation in the clock: $\mathbb{E}[G_t]=t$ and $\operatorname{Var}(G_t)=t/\kappa$. Let $U=G_t$ be the elapsed operational time and $T=G_1$ the total operational time. The ratio $s=U/T$ measures progress from zero to one. Given the clock, the bridge state follows
\begin{equation}
    q(\X_t\mid\X_0,\Y,U,T)
    =\N\!\left((1-s)\X_0+s\Y,
    \sigma_B^2\frac{U(T-U)}{T}\I\right).
    \label{eq:gamma_bridge_family}
\end{equation}
Thus, $s$ controls the position of the mean between the endpoints, while $T$ also affects the noise variance. Normalizing progress does not remove the effect of $T$. Averaging over random clocks gives a mixture that is generally non-Gaussian. The Gamma clock is a L\'evy process. The matrix bridge, however, has dependent increments because it is conditioned on its endpoints. We therefore call it a \emph{L\'evy-driven bridge}. For this homogeneous Gamma clock, $\kappa\rightarrow\infty$ gives $G_t\rightarrow t$ and recovers Eq.~\eqref{eq:brownian_path_background}.

For a solver with $K$ intervals, $k\in\{1,\ldots,K\}$ denotes the step index. Dividing the interval $[0,1]$ equally gives independent clock increments:
\begin{equation}
\begin{aligned}
    \Delta G_k&\sim\operatorname{Gamma}\!\left(
    \kappa/K,\operatorname{rate}=\kappa\right),\\
    \mathbb{E}[\Delta G_k]&=\frac{1}{K},\qquad
    \operatorname{Var}(\Delta G_k)=\frac{1}{\kappa K}.
\end{aligned}
    \label{eq:gamma_clock_steps}
\end{equation}
We call this policy \emph{Random Gamma}. Every step receives the same amount of time on average, but the sampled increments vary. A larger $\kappa$ reduces this variation, while a smaller $\kappa$ allows more uneven progress. The cumulative time after step $k$ is $U_k=\sum_{i=1}^{k}\Delta G_i$. The total time is $T=U_K$, and normalized progress is $s_k=U_k/T$.

\paragraph{Randomness and time allocation.}
Clock variation and early time allocation are separate choices. To change the allocation, we assign positive weights $w_k$ with $\sum_k w_k=1$ and sample independent increments as
\begin{equation}
\begin{aligned}
    \Delta G_k&\sim\operatorname{Gamma}
    (\kappa w_k,\operatorname{rate}=\kappa),\\
    \mathbb{E}[\Delta G_k]&=w_k,\qquad
    \operatorname{Var}(\Delta G_k)=\frac{w_k}{\kappa}.
\end{aligned}
    \label{eq:gamma_allocation_background}
\end{equation}
Section~\ref{sec:method_sb} uses decreasing weights for front-loaded inference. These weights are fixed before sampling. The actual increments remain random and do not have to decrease.

Both allocations give $T\sim\operatorname{Gamma}(\kappa,\operatorname{rate}=\kappa)$, with mean one and variance $1/\kappa$. A sampled total can still differ from one. The weights control how time is allocated across steps. The parameter $\kappa$ controls variation around that allocation, and $\sigma_B$ sets the Gaussian noise scale given the clock. As $\kappa$ increases, the increments approach the chosen weights. A nonuniform profile therefore remains nonuniform in this limit.

\section{Method}
\label{sec:method}

\subsection{Overview}
\label{sec:overview}

Given source points $\mathcal{P}=\{\mathbf{p}_i\}_{i=1}^{N}$ and target points $\mathcal{Q}=\{\mathbf{q}_j\}_{j=1}^{M}$ at the current resolution, we estimate a soft correspondence matrix in $\mathbb{R}^{N\times M}$. A hierarchical backbone provides point features $\F_s,\F_t$ and a coarser level with $N_c$ source points and $M_c$ target points for initialization.

Figure~\ref{fig:framework} shows the inference procedure. We restore a DDIM solution from the coarser level to the current resolution to form the bridge's initial matrix $\X_0$. At each step, the network uses the current matrix, point features, and geometric feedback to predict a target matching matrix. A stochastic matrix-bridge update then combines this prediction with a sampled clock increment to obtain the next state. The initial matrix, current state, and predicted target all have shape $N\times M$. After $K$ endpoint evaluations, we extract correspondences from the final prediction. Random Gamma and Fixed front-loaded Gamma use the same initial matrix, trained network, and update rule. They differ only in how operational time is assigned to each interval.

\subsection{Matrix-Bridge Initialization}
\label{sec:l1}
\label{sec:lifting}

Following Diff-Reg~\cite{wu2024diff}, a DDIM solver~\cite{song2020denoising} produces a soft matching solution $\widehat{\Y}_{\mathrm{DDIM}}^{c}$ at the coarser resolution. We restore this solution to the current resolution using fixed parent assignments $\mathbf{P}_s\in\{0,1\}^{N\times N_c}$ and $\mathbf{P}_t\in\{0,1\}^{M\times M_c}$, obtaining the matrix-bridge starting point:
\begin{equation}
    \X_0=\U(\widehat{\Y}_{\mathrm{DDIM}}^{c})
    =\mathbf{P}_s\widehat{\Y}_{\mathrm{DDIM}}^{c}\mathbf{P}_t^{\top}.
    \label{eq:lifting}
\end{equation}
The operator $\U$ copies assignment values through the parent mapping, so entries within each parent block initially share the same value. The matrix bridge then refines these assignments using point features and the evolving state.

\subsection{Matrix-Bridge Network with Geometry Feedback}
\label{sec:shared_bridge}

Each bridge state provides geometric input for the next endpoint prediction. We first clip the state to $\mathbf{C}_k=\clip(\X_k,0,1)$. We treat $\mathbf{C}_k$ as assignment scores, set invalid scores to $-\infty$, and add a dustbin row and column for unmatched points. We apply Sinkhorn normalization in the log domain, exponentiate the resulting log assignments, and remove the dustbin row and column to obtain nonnegative weights $\mathbf{A}_k$. The $-\infty$ scores thus represent zero assignment weights. For geometric conditioning, Soft Procrustes retains the highest-weight pairs in an index set $\mathcal{S}_k$ and estimates a rigid alignment:
\begin{equation}
    (\mathbf{R}_k,\bm\tau_k)=
    \arg\min_{\mathbf{R}\in\mathrm{SO}(3),\bm\tau}
    \sum_{(i,j)\in\mathcal{S}_k} A_{k,ij}
    \|\mathbf{R}\mathbf{p}_i+\bm\tau-\mathbf{q}_j\|_2^2.
    \label{eq:soft_proc}
\end{equation}
A transformer takes the warped source $\widetilde{\mathbf{p}}_{i,k}=\mathbf{R}_k\mathbf{p}_i+\bm\tau_k$, target geometry, point features, and normalized operational progress $s_k=U_k/T$ as input. The rigid warp provides geometric guidance even for non-rigid correspondence estimation. The matching head predicts the target matching matrix:
\begin{equation}
    \widehat{\Y}_k
    =H_{\phi}\!\left(\mathbf{C}_k,s_k,\F_s,\F_t,
    \widetilde{\mathcal{P}}_k,\mathcal{Q}\right).
    \label{eq:shared_endpoint}
\end{equation}
The clipped matrix also adds a bias to the matching logits. We recompute the alignment at every step and use an identity warp when the Procrustes solution is ill-conditioned. Pair selection affects only the warp; the bridge update retains the full unbounded state $\X_k$.

\subsection{Random Gamma and Fixed Front-Loaded Gamma}
\label{sec:method_sb}

\paragraph{Random Gamma.}
Random Gamma uses $w_k=1/K$ in Eq.~\eqref{eq:gamma_allocation_background}. Each interval receives the same expected operational time. The increments are sampled independently, so progress can be uneven within a run.

\paragraph{Fixed front-loaded Gamma.}

Fixed front-loaded Gamma sets the expected time allocation before sampling the increments. It gives larger weights to early intervals, so they receive more operational time on average. Later endpoint evaluations can then refine the changed state. For $K\geq2$ intervals and $k=1,\ldots,K$, we use a smooth exponential profile with $\tau_k=(k-1)/(K-1)$:
\begin{equation}
\begin{aligned}
    w_k(\beta)&=\frac{\exp(-\beta\tau_k)}
    {\sum_{j=1}^{K}\exp(-\beta\tau_j)},\\
    \Delta G_k&\sim\operatorname{Gamma}
    (\kappa w_k,\operatorname{rate}=\kappa).
\end{aligned}
    \label{eq:frontloaded_clock}
\end{equation}
The strength $\beta\geq0$ controls how quickly the expected allocation decreases. Setting $\beta=0$ recovers Random Gamma. All weights are positive, so later updates still receive operational time. The independent increments share the same rate and satisfy $\sum_k w_k=1$. Their sum therefore follows
\begin{equation}
    \sum_{k=1}^{K}\Delta G_k
    \sim\operatorname{Gamma}(\kappa,\operatorname{rate}=\kappa).
    \label{eq:frontloaded_total}
\end{equation}
Thus, front-loading changes the distribution at each step but keeps the same total-time distribution. We can write the clock as $G_{a(t)}$, where the fixed increasing map $a(t)$ has grid increments $w_k$. For $\beta>0$, equal intervals in solver time have different increment distributions. The increments are therefore independent but nonstationary. This clock is a deterministic time change of the homogeneous Gamma process. We sample each increment directly from its assigned distribution in Eq.~\eqref{eq:frontloaded_clock}; we do not sort the sequence. The trained endpoint network remains unchanged.

\subsection{Bridge Inference}
\label{sec:inference}

Inference starts from the lifted matrix $\X_0$. We sample a clock using Eq.~\eqref{eq:gamma_clock_steps} or Eq.~\eqref{eq:frontloaded_clock}.

For the sampled clock, we set $U_0=0$, $U_k=\sum_{i\leq k}\Delta G_i$, $s_k=U_k/T$, and $R_k=T-U_k$. At step $k$, the network predicts $\widehat{\Y}_k$ from the clipped state, as in Eq.~\eqref{eq:shared_endpoint}. We hold this endpoint fixed for one step. The reference-bridge transition then has mean
\begin{equation}
    \mathbf{m}_{k+1}=\X_k+
    \frac{\Delta G_{k+1}}{R_k}(\widehat{\Y}_k-\X_k),
    \label{eq:gamma_transition_mean}
\end{equation}
with
\begin{equation}
    \begin{aligned}
    V_{k+1}&=\sigma_B^2\frac{\Delta G_{k+1}
    (R_k-\Delta G_{k+1})}{R_k},\\
    \X_{k+1}&=\mathbf{m}_{k+1}+\sqrt{V_{k+1}}\,\mathbf{z}_k,
    \qquad \mathbf{z}_k\sim\N(\mathbf{0},\I).
    \end{aligned}
    \label{eq:gamma_bridge_transition}
\end{equation}
These transitions are exact for the Brownian reference bridge given the clock and a fixed endpoint. The learned solver makes a new endpoint prediction after each step.

The update in Fig.~\ref{fig:framework}(b) combines Eqs.~\eqref{eq:gamma_transition_mean} and~\eqref{eq:gamma_bridge_transition}, with $\alpha_k=\Delta G_{k+1}/R_k$. This coefficient controls how far the mean moves toward the current endpoint prediction. For the same state, endpoint, and remaining time, a larger increment gives a larger mean change. The variance also depends on the time left after the update. It becomes zero when all remaining time is used. Thus, the clock increment alone does not determine the actual matrix change. The endpoint estimate and Gaussian noise also matter. The updated state then provides the geometry and matching bias for the next prediction.

We apply stochastic transitions for the first $K-1$ increments and return $\widehat{\Y}_{K-1}$ from the last state. Using the final increment would set the mean to this endpoint and the variance to zero. No further stochastic update is therefore needed. The matrix bridge uses $K$ endpoint-network evaluations. We clip the final endpoint matrix to $[0,1]$ and extract correspondences by mutual matching and confidence filtering.

\section{Experiments}
\label{sec:experiments}

\subsection{Evaluation Protocol}

We evaluate non-rigid correspondence on 4DMatch and 4DLoMatch following Lepard~\cite{li2022lepard}. We use non-rigid feature matching recall (NFMR) and inlier ratio (IR). NFMR is the fraction of benchmark test points whose interpolated motion has endpoint error below 0.04. IR is the fraction of extracted matches whose error under the ground-truth deformation is below 0.04. Unless otherwise specified, comparisons within our model use the same backbone, preprocessing, checkpoint, lifted initial matrix, and $K=20$ endpoint-network evaluations. We set $\kappa=10$ for Gamma clocks and $\sigma_B=0.1$ for stochastic bridge updates. We use five DDIM evaluations to construct the coarse matching solution and a confidence threshold of $0.2$ for final mutual matching.

Fixed front-loaded Gamma uses $\beta=2$ in the main comparison and visualizations. We also test $\beta\in\{0,0.5,1,2,4\}$ on both evaluation benchmarks to study the effect of strength. Multi-seed results report mean$\pm$population standard deviation over inference seeds 0, 1, and 2 with one checkpoint. We first examine controls for the clock, noise, initial matrix, and increment order. We then compare front-loaded policies and their inference cost.

\begin{table*}[t]
\caption{Correspondence results on 4DMatch and 4DLoMatch (\%). Baseline values are from the cited papers. Both L\'evyMatch policies use the same checkpoint. L\'evyMatch rows report mean$\pm$population standard deviation over three inference seeds. Fixed front-loaded Gamma uses $\beta=2$.}
\label{tab:main_results}
\centering
\resizebox{0.7\linewidth}{!}{
\begin{tabular}{lcccc}
\toprule
& \multicolumn{2}{c}{4DMatch} & \multicolumn{2}{c}{4DLoMatch}\\
\cmidrule(lr){2-3}\cmidrule(lr){4-5}
Method & NFMR$\uparrow$ & IR$\uparrow$ & NFMR$\uparrow$ & IR$\uparrow$\\
\midrule
PointPWC~\cite{wu2019pointpwc} & 21.60 & 20.00 & 10.00 & 7.20\\
FLOT~\cite{puy2020flot} & 27.10 & 24.90 & 15.20 & 10.70\\
D3Feat~\cite{bai2020d3feat} & 55.50 & 54.70 & 27.40 & 21.50\\
Predator~\cite{huang2021predator} & 56.40 & 60.40 & 32.10 & 27.50\\
Lepard~\cite{li2022lepard} & 83.60 & 82.64 & 66.63 & 55.55\\
GeoTransformer~\cite{qin2022geometric} & 83.20 & 82.20 & 65.40 & 63.60\\
RoITr~\cite{yu2023rotation} & 83.00 & 84.40 & 69.40 & 67.60\\
Diff-Reg~\cite{wu2024diff} & 90.25 & 87.98 & 77.15 & 67.00\\
\midrule
L\'evyMatch, Random Gamma & $92.88{\pm}0.08$ & $91.95{\pm}0.06$ & $82.32{\pm}0.18$ & $78.61{\pm}0.17$\\
L\'evyMatch, Fixed front-loaded & $\mathbf{93.09{\pm}0.07}$ & $\mathbf{92.11{\pm}0.05}$ & $\mathbf{82.79{\pm}0.19}$ & $\mathbf{79.07{\pm}0.18}$\\
\bottomrule
\end{tabular}}
\end{table*}

\begin{table*}[t]
\caption{Inference-clock and noise controls. All rows use the same checkpoint and endpoint-network evaluation budget. Only the inference clock or transition noise changes.}
\label{tab:gamma_ablation}
\centering
\resizebox{0.7\textwidth}{!}{
\begin{tabular}{llcccc}
\toprule
& & \multicolumn{2}{c}{4DMatch} & \multicolumn{2}{c}{4DLoMatch}\\
\cmidrule(lr){3-4}\cmidrule(lr){5-6}
Variant & Clock & NFMR$\uparrow$ & IR$\uparrow$ & NFMR$\uparrow$ & IR$\uparrow$\\
\midrule
Full checkpoint, Random/SDE & Gamma & $92.88{\pm}0.08$ & $91.95{\pm}0.06$ & $82.32{\pm}0.18$ & $78.61{\pm}0.17$\\
Full checkpoint, uniform/SDE & uniform & $92.82{\pm}0.09$ & $91.89{\pm}0.07$ & $82.32{\pm}0.34$ & $78.61{\pm}0.32$\\
Full checkpoint, uniform/ODE & uniform & $92.85{\pm}0.05$ & $91.91{\pm}0.06$ & $82.21{\pm}0.23$ & $78.47{\pm}0.18$\\
\bottomrule
\end{tabular}}
\end{table*}

\begin{table*}[t]
\caption{Initial-matrix and bridge-bypass controls using the same checkpoint. Rows with the bridge report results over three inference seeds. The bypass row uses seed 0.}
\label{tab:init_ablation}
\centering
\resizebox{0.82\textwidth}{!}{
\begin{tabular}{llcccc}
\toprule
& & \multicolumn{2}{c}{4DMatch} & \multicolumn{2}{c}{4DLoMatch}\\
\cmidrule(lr){3-4}\cmidrule(lr){5-6}
Initialization & L\'evy matrix bridge & NFMR$\uparrow$ & IR$\uparrow$ & NFMR$\uparrow$ & IR$\uparrow$\\
\midrule
Random valid matrix & \checkmark & $92.30{\pm}0.12$ & $91.46{\pm}0.09$ & $\mathbf{82.94{\pm}0.09}$ & $\mathbf{79.39{\pm}0.10}$\\
Lifted DDIM-5 & -- & 48.54 & 36.06 & 34.79 & 18.13\\
Lifted DDIM-5 & \checkmark & $\mathbf{92.88{\pm}0.08}$ & $\mathbf{91.95{\pm}0.06}$ & $82.32{\pm}0.18$ & $78.61{\pm}0.17$\\
\bottomrule
\end{tabular}}
\end{table*}

\begin{table*}[t]
\caption{Increment-order diagnostic. Gamma rows use the same sampled increment values for each seed and report three-seed results. The uniform row uses deterministic mean updates and seed 0. $\Delta$Entropy and $\Delta$Geo. are final-minus-initial changes in mean row entropy (nats) and weighted mean Euclidean alignment residual (point-coordinate units), respectively. Their definitions are given in Eqs.~\eqref{eq:diagnostic_entropy}--\eqref{eq:diagnostic_changes}. Matching entropy does not measure accuracy.}
\label{tab:jump_timing}
\centering
\small
\setlength{\tabcolsep}{8pt}
\begin{tabular}{lcccc}
\toprule
\multicolumn{5}{c}{4DMatch}\\
\midrule
Clock order & NFMR$\uparrow$ & IR$\uparrow$ & $\Delta$Entropy & $\Delta$Geo.$\downarrow$\\
\midrule
Uniform/ODE (seed 0) & 92.79 & 91.85 & -0.5294 & -0.0318\\
Gamma, random & $92.88{\pm}0.08$ & $91.95{\pm}0.06$ & $-0.5291{\pm}0.0006$ & $-0.0313{\pm}0.0002$\\
Gamma, large early & $\mathbf{93.23{\pm}0.09}$ & $\mathbf{92.23{\pm}0.06}$ & $-0.5402{\pm}0.0005$ & $\mathbf{-0.0322{\pm}0.0001}$\\
Gamma, large late & $92.23{\pm}0.14$ & $91.41{\pm}0.13$ & $-0.5040{\pm}0.0007$ & $-0.0296{\pm}0.0002$\\
\midrule
\multicolumn{5}{c}{4DLoMatch}\\
\midrule
Clock order & NFMR$\uparrow$ & IR$\uparrow$ & $\Delta$Entropy & $\Delta$Geo.$\downarrow$\\
\midrule
Uniform/ODE (seed 0) & 82.02 & 78.32 & 0.6067 & -0.0044\\
Gamma, random & $82.32{\pm}0.18$ & $78.61{\pm}0.17$ & $0.6134{\pm}0.0031$ & $-0.0040{\pm}0.0004$\\
Gamma, large early & $\mathbf{82.80{\pm}0.25}$ & $\mathbf{79.03{\pm}0.23}$ & $0.6116{\pm}0.0029$ & $\mathbf{-0.0057{\pm}0.0003}$\\
Gamma, large late & $81.08{\pm}0.12$ & $77.51{\pm}0.14$ & $0.6123{\pm}0.0028$ & $0.0004{\pm}0.0003$\\
\bottomrule
\end{tabular}
\end{table*}

\begin{table*}[t]
\caption{Effect of front-loading strength over three inference seeds. All settings use the same checkpoint. Increments are sampled from the assigned Gamma distributions without sorting.}
\label{tab:fixed_frontloaded}
\centering
\resizebox{0.76\textwidth}{!}{
\begin{tabular}{lccccc}
\toprule
& & \multicolumn{2}{c}{4DMatch} & \multicolumn{2}{c}{4DLoMatch}\\
\cmidrule(lr){3-4}\cmidrule(lr){5-6}
Clock & \(\beta\) & NFMR\(\uparrow\) & IR\(\uparrow\) & NFMR\(\uparrow\) & IR\(\uparrow\)\\
\midrule
Random Gamma & 0 & $92.88{\pm}0.08$ & $91.95{\pm}0.06$ & $82.32{\pm}0.18$ & $78.61{\pm}0.17$\\
Fixed front-loaded & 0.5 & $92.87{\pm}0.10$ & $91.93{\pm}0.09$ & $82.58{\pm}0.17$ & $78.86{\pm}0.16$\\
Fixed front-loaded & 1.0 & $92.96{\pm}0.09$ & $92.00{\pm}0.07$ & $82.63{\pm}0.08$ & $78.92{\pm}0.05$\\
Fixed front-loaded & 2.0 & $93.09{\pm}0.07$ & $92.11{\pm}0.05$ & $82.79{\pm}0.19$ & $79.07{\pm}0.18$\\
Fixed front-loaded & 4.0 & $\mathbf{93.17{\pm}0.07}$ & $\mathbf{92.19{\pm}0.05}$ & $\mathbf{82.98{\pm}0.26}$ & $\mathbf{79.21{\pm}0.23}$\\
\bottomrule
\end{tabular}}
\end{table*}

\begin{table*}[t]
\caption{Recursive inference cost and output coverage for the main comparison. Fixed front-loaded Gamma uses $\beta=2$. Both policies use 20 endpoint-network evaluations. Values are means over inference seeds 0, 1, and 2. Time includes the backbone, coarse DDIM source, and matrix-bridge solver, but excludes the downstream deformation solver. Matches is the number of retained correspondences per pair. Empty is the percentage of pairs with no retained matches.}
\label{tab:inference_cost}
\centering
\resizebox{0.7\textwidth}{!}{
\begin{tabular}{llccc}
\toprule
Dataset & Clock & Time (ms)$\downarrow$ & Matches & Empty (\%)$\downarrow$\\
\midrule
4DMatch & Random Gamma & 479.8 & 648.82 & 0.00\\
4DMatch & Fixed front-loaded ($\beta=2$) & 500.9 & 650.33 & 0.00\\
\addlinespace
4DLoMatch & Random Gamma & 493.2 & 379.99 & 0.00\\
4DLoMatch & Fixed front-loaded ($\beta=2$) & 481.8 & 380.18 & 0.00\\
\bottomrule
\end{tabular}}
\end{table*}

\begin{figure*}[t]
\centering
\includegraphics[width=0.7\textwidth]{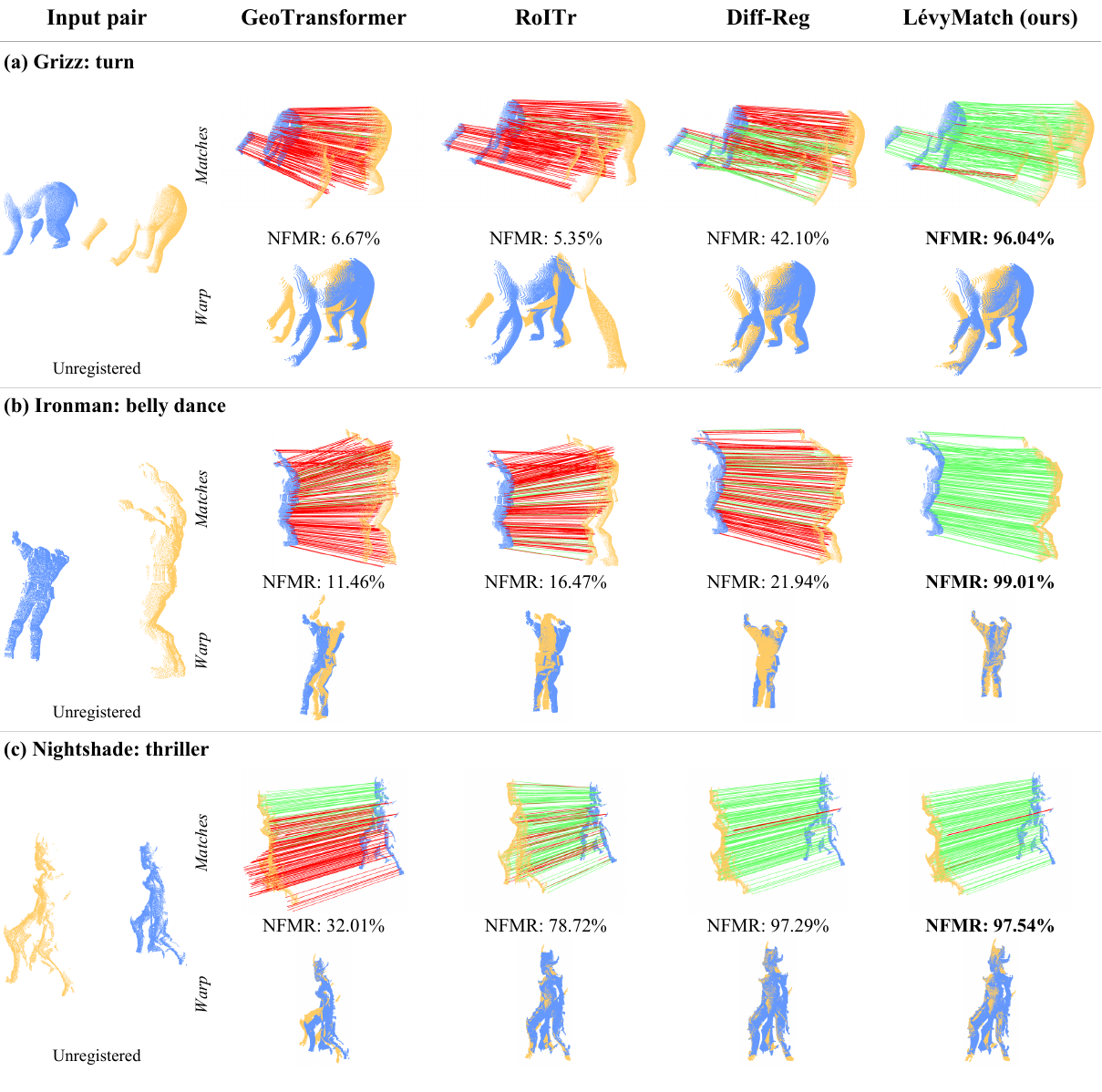}
\caption{Selected qualitative comparisons on 4DMatch. L\'evyMatch uses Fixed front-loaded Gamma with $\beta=2$ and inference seed 0. For each pair, the left column shows the inputs before registration. The upper row shows interpolated test-point correspondences. The lower row shows the warped source (yellow) and target (blue). Green lines mark endpoint errors below 0.04 in dataset coordinate units; red lines mark errors at or above this threshold. The clouds are shown apart in the input and correspondence views. Camera views differ across panels.}
\label{fig:qualitative_4dmatch}
\end{figure*}

\begin{figure*}[t]
\centering
\includegraphics[width=0.7\textwidth]{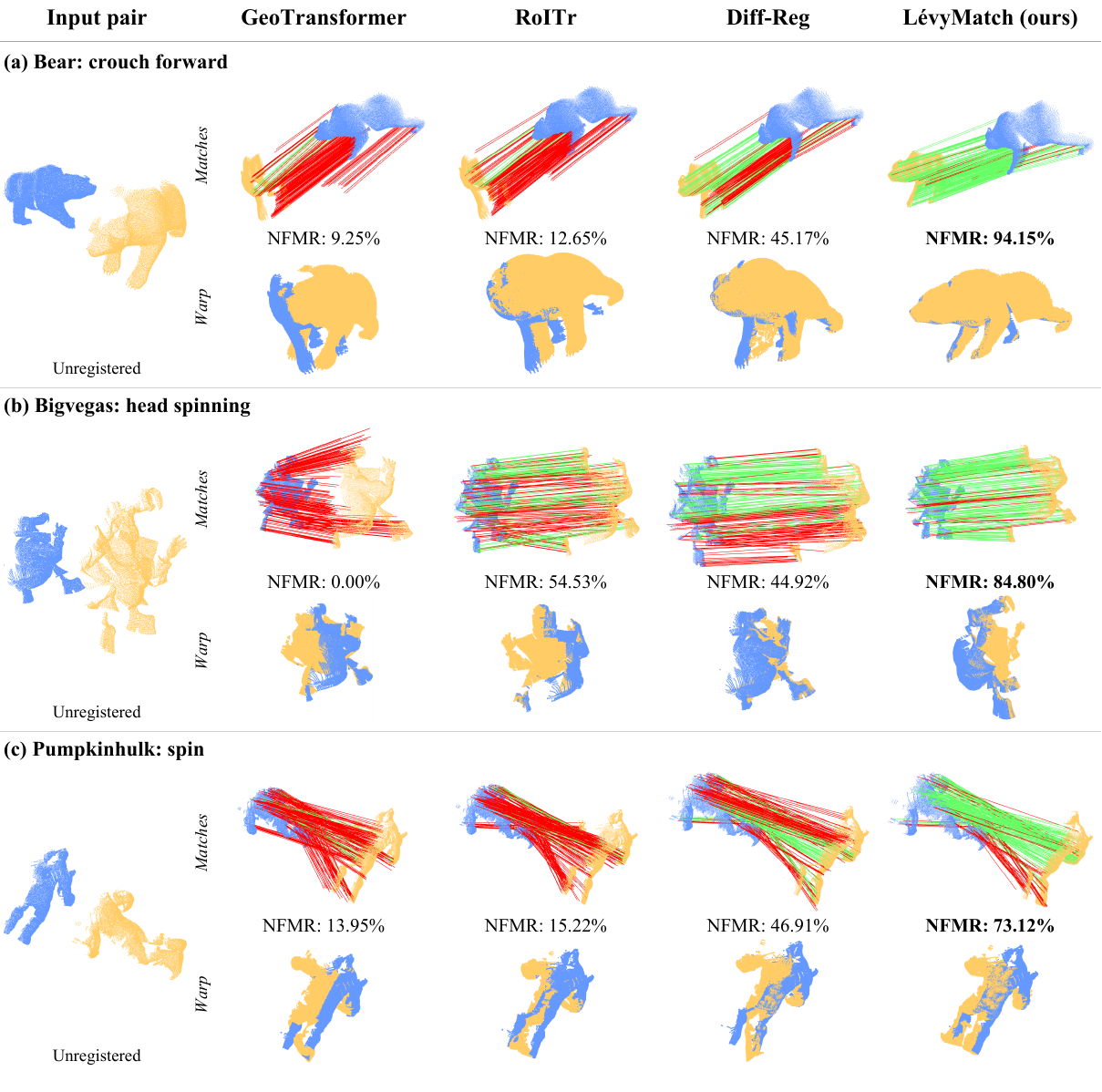}
\caption{Selected qualitative comparisons on 4DLoMatch, using the layout and correctness threshold of Fig.~\ref{fig:qualitative_4dmatch}. L\'evyMatch uses Fixed front-loaded Gamma with $\beta=2$ and inference seed 0. The Pumpkinhulk example still has incorrect correspondences and visible warp errors around the extended limbs.}
\label{fig:qualitative_4dlomatch}
\end{figure*}

\subsection{Main Correspondence Results}

Table~\ref{tab:main_results} compares correspondence accuracy. Random Gamma reaches $92.88\pm0.08$ NFMR on 4DMatch and $82.32\pm0.18$ on 4DLoMatch. The default front-loaded clock raises these values to $93.09\pm0.07$ and $82.79\pm0.19$. IR reaches $92.11\pm0.05$ and $79.07\pm0.18$, respectively. Thus, front-loading gives accurate motion estimates for more test points and increases the fraction of correct extracted matches. The gain is larger on the low-overlap benchmark.
Compared with Diff-Reg, the front-loaded model improves NFMR/IR by 2.84/4.13 points on 4DMatch and 5.64/12.07 points on 4DLoMatch. The following experiments examine inference-clock choices within \paper using the same trained weights and matched settings.

\subsection{Random Gamma Bridge Analysis}

Table~\ref{tab:gamma_ablation} compares Random Gamma with two uniform-clock controls. All three rows use the same checkpoint, initial-matrix construction, and endpoint-evaluation budget. The uniform/SDE control sets $\Delta G_k=1/K$ and keeps Gaussian noise in the bridge transition. The uniform/ODE control uses the same fixed clock but removes this noise. Here, ``uniform/ODE'' denotes deterministic matrix-bridge mean updates. These comparisons test the effects of inference-clock randomness and transition noise.

Random timing alone has little effect on average accuracy in this comparison. Random Gamma and uniform/SDE differ by only about 0.06 NFMR points on 4DMatch and have nearly identical means on 4DLoMatch. Removing transition noise from the uniform updates also has a small, dataset-dependent effect: NFMR rises by about 0.02 points on 4DMatch and falls by 0.12 points on 4DLoMatch. Changing from fixed to random time increments gives little improvement here. We therefore also test whether using the same large increments earlier or later changes the result.

\paragraph{Structured initialization.}
Table~\ref{tab:init_ablation} replaces the lifted DDIM-5 matrix with independent $\operatorname{Uniform}(0,1)$ values on valid entries, or evaluates the lifted matrix without matrix-bridge refinement. Both refined variants use Random Gamma and the same point coordinates, backbone features, and endpoint network.

Removing the refinement stage reduces accuracy much more than replacing its initial matrix. Without refinement, the lifted DDIM solution reaches only 48.54/34.79 NFMR on 4DMatch/4DLoMatch, compared with 92.88/82.32 after refinement. Starting from a random matrix still gives 92.30/82.94. Thus, accurate correspondences do not simply come from copying the coarse solution: the endpoint network, geometric feedback, and repeated updates together provide substantial refinement. The lifted source helps on 4DMatch, while the random source performs better on 4DLoMatch, so neither initial matrix gives the best result on both datasets. In both cases, the network produces accurate matches after refinement.

\paragraph{Gamma-increment order.}
For each seed, we use the same sampled Gamma increments in three orders: random, descending (large early), and ascending (large late). Sorting keeps the increment values and total time fixed but changes when the increments are used. Table~\ref{tab:jump_timing} reports correspondence accuracy and includes a single-seed uniform-clock reference. The same table reports matching-entropy and geometric-residual changes.

The ordering results show that the same increment values can lead to different final accuracy. Early placement performs best on both datasets, followed by random and then late placement. On 4DLoMatch, early placement raises NFMR by about 0.48 points over random ordering, whereas late placement lowers it by about 1.24 points. The same asymmetry appears on 4DMatch, with a gain of about 0.34 points and a loss of about 0.65 points. Placing large increments late therefore hurts more than placing them early helps. One possible explanation is that, after an early update, the network still has several steps to recompute the alignment and revise its predicted matches. After a late update, fewer such steps remain. The last increment also matters: it is reserved for endpoint prediction, so placing the largest increment last makes this prediction occur earlier in normalized bridge time.

\paragraph{Diagnostic measures.}
For each matrix $\mathbf{M}$, both diagnostics use clipped weights $W_{ij}=\clip(M_{ij},0,1)$ on valid entries and zero elsewhere. We compute them before mutual correspondence extraction, without extra Sinkhorn normalization. Let $r_i=\sum_j W_{ij}$, $p_{ij}=W_{ij}/\max(r_i,\epsilon)$, and $\mathcal{I}(\mathbf{M})=\{i:r_i>\epsilon\}$, with $\epsilon=10^{-12}$. Matching entropy is the mean row entropy over rows whose total weight is nonzero:
\begin{equation}
    H(\mathbf{M})=
    -\frac{\displaystyle\sum_{i\in\mathcal{I}(\mathbf{M})}\sum_j
    p_{ij}\log\max(p_{ij},\epsilon)}
    {\max(1,|\mathcal{I}(\mathbf{M})|)}.
    \label{eq:diagnostic_entropy}
\end{equation}
Lower row entropy means that a source point assigns most of its weight to fewer target points; it does not tell us whether those targets are correct. We use natural logarithms, so entropy is measured in nats. We do not normalize this value by the number of columns. It differs from the normalized entropy of the clock profile.

For the geometric diagnostic, let $(\mathbf{R}_{\mathbf{M}},\bm\tau_{\mathbf{M}})$ be the alignment returned by soft Procrustes using $\mathbf{W}$. We measure the weighted mean Euclidean residual
\begin{equation}
    D_{\mathrm{geo}}(\mathbf{M})=
    \frac{\displaystyle\sum_{ij}W_{ij}
    \|\mathbf{R}_{\mathbf{M}}\mathbf{p}_i+
    \bm\tau_{\mathbf{M}}-\mathbf{q}_j\|_2}
    {\max(\epsilon,\sum_{ij}W_{ij})}.
    \label{eq:diagnostic_geo}
\end{equation}
A smaller residual means that point pairs with high matching weights lie closer together after the estimated rigid alignment. It measures agreement with that alignment, not error against ground-truth correspondences. The distances are not squared, so the residual has the same units as the point coordinates. For each test pair, we compute
\begin{equation}
    \begin{aligned}
    \Delta\mathrm{Entropy}&=H(\widehat{\Y}^{h})-H(\X_0),\\
    \Delta\mathrm{Geo.}&=D_{\mathrm{geo}}(\widehat{\Y}^{h})-D_{\mathrm{geo}}(\X_0),
    \end{aligned}
    \label{eq:diagnostic_changes}
\end{equation}
where $\X_0$ is the lifted initial matrix and $\widehat{\Y}^{h}$ is the final predicted matching matrix. We first average these changes over test pairs. For the Gamma rows, we then report the mean and population standard deviation across inference seeds 0, 1, and 2. The uniform reference uses only seed 0. A negative value means that the diagnostic decreases. It does not necessarily mean that correspondence accuracy improves.

\paragraph{Alignment residual and matching entropy.}
Early ordering gives both higher accuracy and a larger reduction in the alignment residual (Table~\ref{tab:jump_timing}). On 4DLoMatch, early placement reduces the geometric residual by 0.0057, while late placement increases it by 0.0004. The residual also improves more with early placement on 4DMatch. Matching entropy, however, decreases on 4DMatch and increases on 4DLoMatch for all tested orders. On 4DLoMatch, early and late placement have similar entropy changes despite their different accuracy. Thus, early ordering gives higher NFMR and IR and a lower weighted alignment residual, but does not always make the matching weights more concentrated. Concentration alone cannot explain why early ordering is more accurate than late ordering.

\subsection{Fixed Front-Loaded Clock}

Table~\ref{tab:fixed_frontloaded} tests the exponential shape profile in Eq.~\eqref{eq:frontloaded_clock}. Setting $\beta=0$ gives Random Gamma. Positive values assign more expected time to early steps. All tested strengths use the same checkpoint and total-time distribution. Unlike the ordering diagnostic, each front-loaded policy samples directly from the Gamma distribution assigned to each step.

Fixed front-loaded Gamma samples larger increments on average at early steps, without sorting them after sampling. At the default $\beta=2$, NFMR improves over Random Gamma by about 0.20 points on 4DMatch and 0.47 points on 4DLoMatch, and both mean NFMR and mean IR improve. The mean number of retained matches changes only from 648.82 to 650.33 and from 379.99 to 380.18, respectively. The method returns almost the same number of matches, but a larger fraction is correct and more test points have accurate motion estimates. The gain therefore cannot be explained simply by returning more matches. Table~\ref{tab:inference_cost} reports the mean output counts and runtime.

The effect of strength differs between the datasets. On 4DLoMatch, every tested positive $\beta$ improves both mean metrics. On 4DMatch, $\beta=0.5$ performs similarly to Random Gamma, while higher strengths improve the means. The largest tested value, $\beta=4$, gives the highest means on both datasets. Compared with $\beta=2$, it adds 0.084 and 0.184 NFMR points. The corresponding standard deviation across inference seeds on 4DLoMatch increases from 0.19 to 0.26 points.

Direct front-loaded sampling achieves accuracy close to early sorting on 4DLoMatch: 82.79/79.07 NFMR/IR versus 82.80/79.03. Early sorting remains slightly better on 4DMatch. The two policies use different clock distributions: sorting preserves each sampled set, while the fixed profile changes the increment distributions and keeps the same total-time distribution. Both improve over Random Gamma. Thus, the tested model can benefit from larger expected increments at early steps without first sampling and sorting an entire sequence.

The experiments show a clear difference between adding random timing and choosing when larger increments occur. Random Gamma gives accuracy close to uniform stochastic updates. Early sorting and Fixed front-loaded Gamma improve accuracy while keeping the same number of endpoint predictions. Early sorting also reduces the alignment residual more, and the fixed policy improves accuracy with almost unchanged match counts. These results support using larger expected time increments early, followed by smaller expected increments for the remaining updates.

\paragraph{Inference cost.}
Both policies use 20 endpoint-network evaluations (19 stochastic transitions and a final endpoint prediction), with mean runtimes of approximately 480--501 ms per pair across the two benchmarks. These measurements include the backbone, coarse DDIM source, and matrix-bridge solver, excluding the downstream deformation solver; Table~\ref{tab:inference_cost} reports the timing and output-coverage results.

\subsection{Qualitative Results}
\label{sec:qualitative}

Figures~\ref{fig:qualitative_4dmatch} and~\ref{fig:qualitative_4dlomatch} show interpolated test-point predictions and the resulting non-rigid warps from the GraphSCNet~\cite{qin2023deep} evaluation pipeline. We use $\beta=2$ and inference seed 0. We selected six pairs with good \paper performance to show individual results. NFMR uses all test points in each pair, while each correspondence panel shows at most 200 predictions. The warped point clouds show how the estimated correspondences affect the final alignment.

On 4DMatch, \paper reaches 96.04\% and 99.01\% NFMR on the Grizz turn and Ironman belly-dance pairs, compared with 42.10\% and 21.94\% for Diff-Reg. In the Ironman example, correct predictions cover the torso and limbs. The warped source follows the raised arms and body. The methods perform similarly on the Nightshade pair, with mostly correct predictions and similar warps. NFMR is 97.29\% for Diff-Reg and 97.54\% for \paper.

On 4DLoMatch, the Bear, Bigvegas, and Pumpkinhulk pairs reach NFMR of 94.15\%, 84.80\%, and 73.12\%, respectively. Correct predictions cover more of the displayed source points, but errors remain under partial overlap. Pumpkinhulk improves over Diff-Reg's 46.91\% NFMR. Incorrect correspondences and visible alignment errors remain around the extended limbs.

\section{Conclusion}
\label{sec:conclusion}

We presented \paper for non-rigid correspondence estimation. Starting from a DDIM solution restored to the current resolution, the method repeatedly predicts a target matching matrix from the current assignments and geometric feedback. A Gamma-clock matrix bridge then updates the assignments toward that prediction. Experiments show that the placement of clock increments matters: using larger increments early gives better results than placing the same increments late, while Random Gamma performs similarly to uniform stochastic updates. The fixed front-loaded policy achieves 93.09\% NFMR and 92.11\% IR on 4DMatch, and 82.79\% NFMR and 79.07\% IR on 4DLoMatch, using the same checkpoint and number of network evaluations as Random Gamma. These results support allocating more clock time to early updates when only a fixed number of refinement steps is available.

\FloatBarrier
{\small
\bibliographystyle{ieeenat_fullname}
\bibliography{main}
}
\end{document}